\documentclass[sigconf]{acmart}

\AtBeginDocument{
\settopmatter{printacmref=false} 
\renewcommand\footnotetextcopyrightpermission[1]{} 
\pagestyle{plain}\pagenumbering{gobble} 
}

\usepackage[draft=false]{defs-common}
\usepackage{defs-custom}

\AtBeginDocument{%
  \providecommand\BibTeX{{%
    \normalfont B\kern-0.5em{\scshape i\kern-0.25em b}\kern-0.8em\TeX}}}

\copyrightyear{2020}
\acmYear{2020}
\setcopyright{acmcopyright}\acmConference[FDG '20]{International Conference on the Foundations of Digital Games}{September 15--18, 2020}{Bugibba, Malta}
\acmBooktitle{International Conference on the Foundations of Digital Games (FDG '20), September 15--18, 2020, Bugibba, Malta}
\acmPrice{15.00}
\acmDOI{10.1145/3402942.3409604}
\acmISBN{978-1-4503-8807-8/20/09}

\begin{document}

\title{Sequential Segment-based Level Generation and Blending using Variational Autoencoders}


\author{Anurag Sarkar}
\affiliation{%
  \institution{Northeastern University}}
\email{sarkar.an@northeastern.edu}

\author{Seth Cooper}
\affiliation{%
  \institution{Northeastern University}}
\email{se.cooper@northeastern.edu}


\begin{abstract}
Existing methods of level generation using latent variable models such as VAEs and GANs do so in segments and produce the final level by stitching these separately generated segments together. In this paper, we build on these methods by training VAEs to learn a sequential model of segment generation such that generated segments logically follow from prior segments. By further combining the VAE with a classifier that determines whether to place the generated segment to the top, bottom, left or right of the previous segment, we obtain a pipeline that enables the generation of arbitrarily long levels that progress in any of these four directions and are composed of segments that logically follow one another. In addition to generating more coherent levels of non-fixed length, this method also enables implicit blending of levels from separate games that do not have similar orientation. We demonstrate our approach using levels from \textit{Super Mario Bros.}, \textit{Kid Icarus} and \textit{Mega Man}, showing that our method produces levels that are more coherent than previous latent variable-based approaches and are capable of blending levels across games.
\end{abstract}

\ccsdesc[300]{Applied computing~Computer games}
\keywords{procedural content generation, variational autoencoder, level generation, level blending, game blending, PCGML}

\maketitle


\newcommand{\XFIGUREblending}{
\begin{figure}[h]
\centering
\includegraphics[width=1\columnwidth]{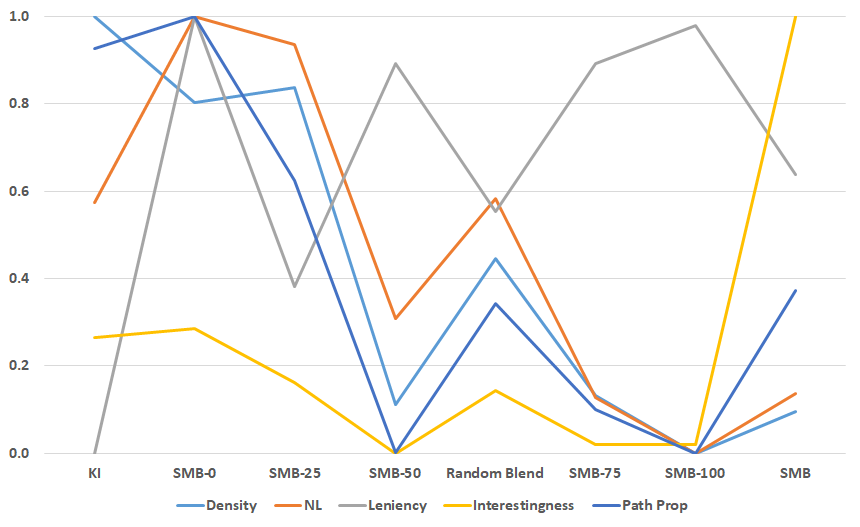}
\caption{\label{XFIGUREblending} Per-segment tile metrics for original SMB and KI levels along with different types of blends. Due to differing scales, all metrics were normalized to be between 0 and 1.}
\end{figure}
}

\newcommand{\XFIGUREblendingsepaltsub}{
\begin{figure*}[t!]
\centering
\begin{subfigure}[t]{0.333\textwidth}
\centering
\includegraphics[width=1\linewidth]{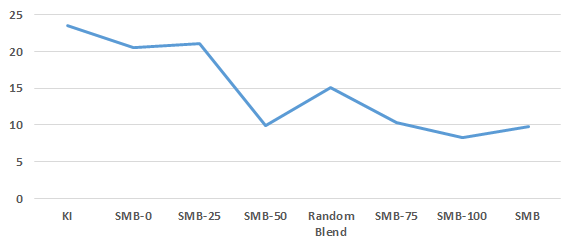}
\caption{Density}
\end{subfigure}
~
\begin{subfigure}[t]{0.333\textwidth}
\centering
\includegraphics[width=1\linewidth]{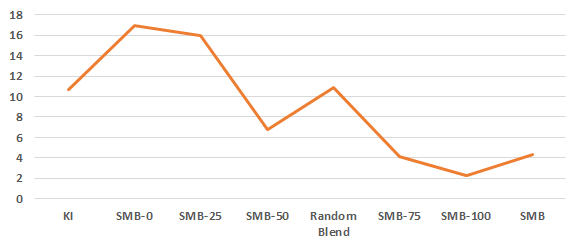}
\caption{Non-Linearity}
\end{subfigure}
~
\begin{subfigure}[t]{0.333\textwidth}
\includegraphics[width=1\linewidth]{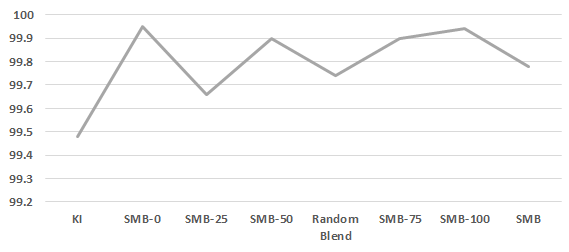}
\caption{Leniency}
\end{subfigure}
\newline
\begin{subfigure}{0.333\textwidth}
\includegraphics[width=1\linewidth]{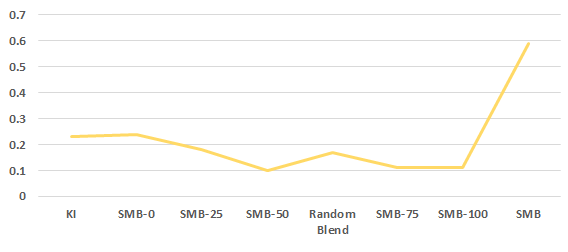}
\caption{Interestingness}
\end{subfigure}
~
\begin{subfigure}{0.333\textwidth}
\includegraphics[width=1\linewidth]{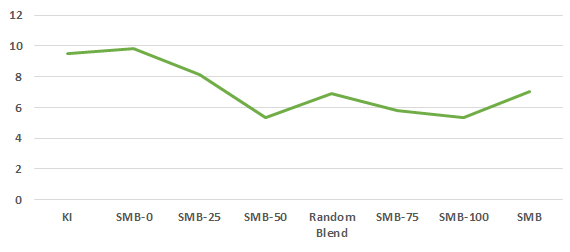}
\caption{Path-Prop}
\end{subfigure}
\caption{\label{XFIGUREblendingsepaltsub} Per-segment tile metrics for original SMB and KI levels along with different types of blends.}
\end{figure*}
}

\newcommand{\XFIGUREprogressionsub}{
\begin{figure*}[t!]
\centering
\begin{subfigure}[t]{0.333\textwidth}
\centering
\includegraphics[width=0.9\linewidth]{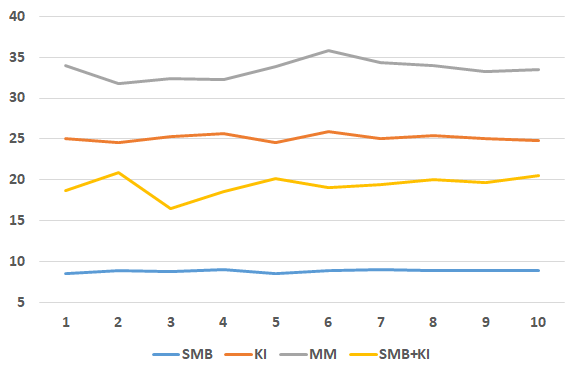}
\caption{Density}
\end{subfigure}
~
\begin{subfigure}[t]{0.333\textwidth}
\centering
\includegraphics[width=0.9\linewidth]{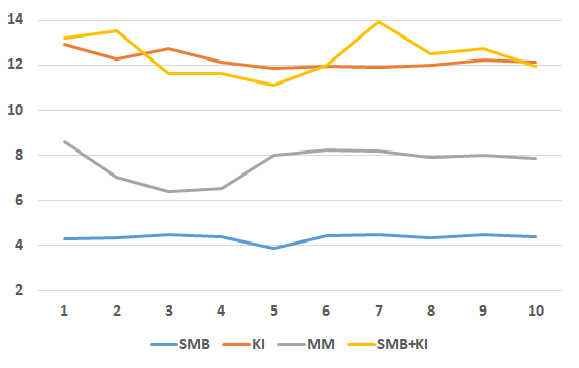}
\caption{Non-Linearity}
\end{subfigure}
~
\begin{subfigure}[t]{0.333\textwidth}
\includegraphics[width=0.9\linewidth]{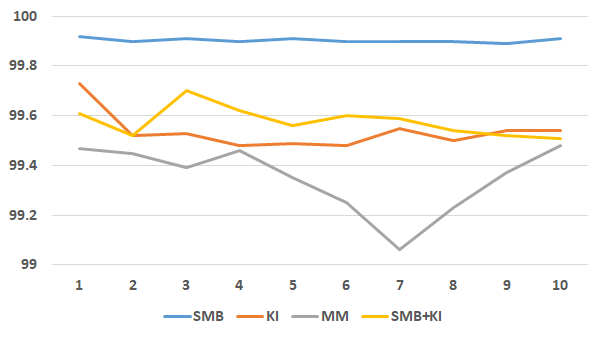}
\caption{Leniency}
\end{subfigure}
\newline
\begin{subfigure}{0.333\textwidth}
\includegraphics[width=0.9\linewidth]{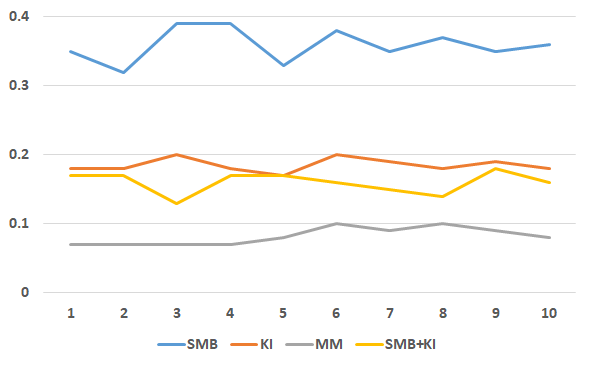}
\caption{Interestingness}
\end{subfigure}
~
\begin{subfigure}{0.333\textwidth}
\includegraphics[width=0.9\linewidth]{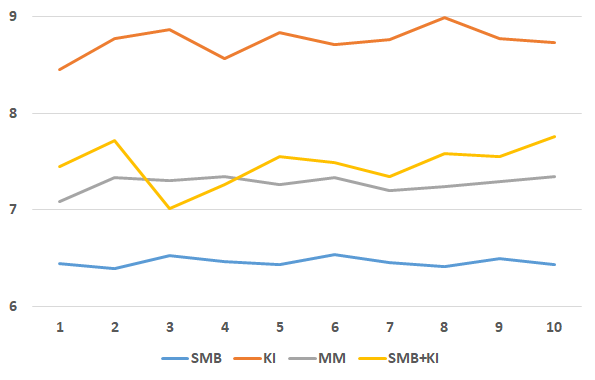}
\caption{Path-Prop}
\end{subfigure}
~
\begin{subfigure}{0.333\textwidth}
\includegraphics[width=0.9\linewidth]{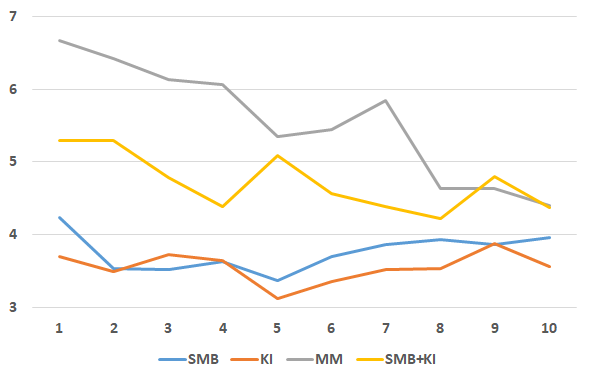}
\caption{Discontinuity}
\end{subfigure}
\caption{\label{XFIGUREprogressionsub} Per-segment metric values plotted for each grouping of 16 segments for MM and each grouping of 12 segments for the other games. x-axis values indicate 1st such grouping, 2nd such grouping etc. y-axis indicates average metric value for the corresponding group of segments.}
\end{figure*}
}

\newcommand{\XFIGURErecon}{
\begin{figure}[t]
\centering
\includegraphics[width=0.975\columnwidth]{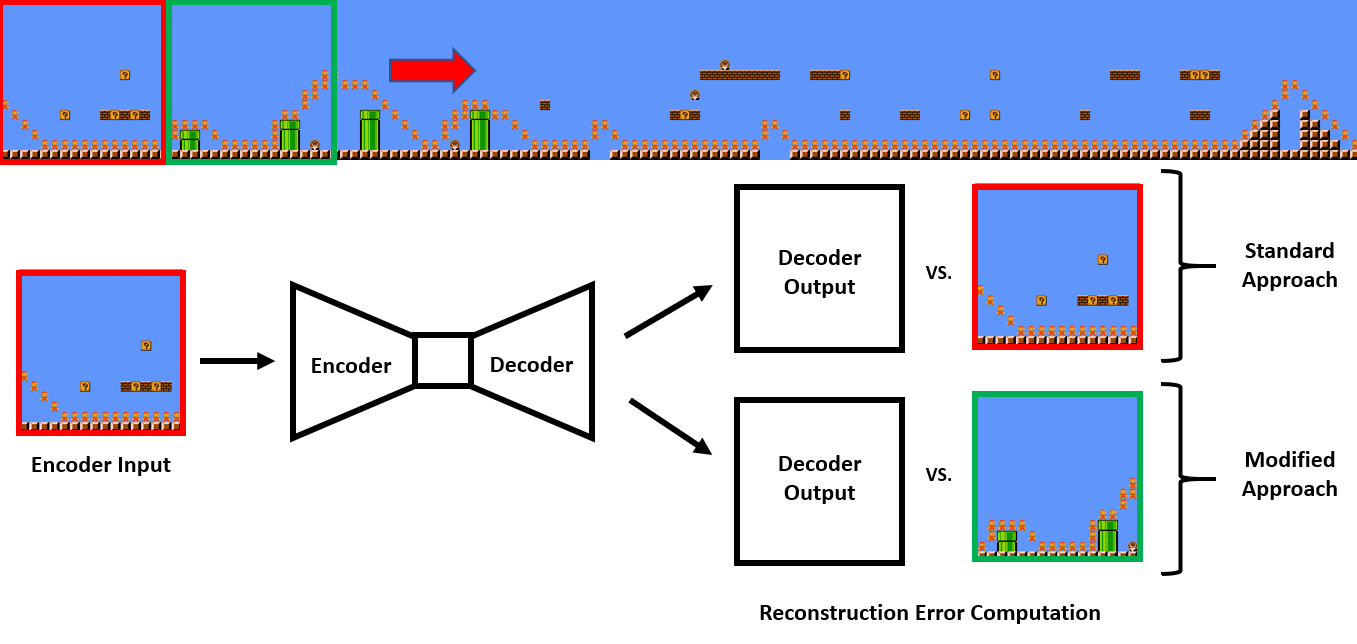}
\caption{\label{XFIGURErecon} Comparison of reconstruction error computation between the standard approach and our modified approach.}
\end{figure}
}

\newcommand{\XFIGUREprogression}{
\begin{figure*}[h]
\centering
\begin{tabular}{ccc}
\includegraphics[width=0.65\columnwidth]{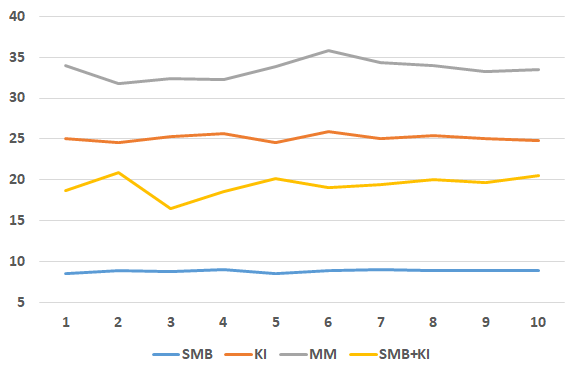} &
\includegraphics[width=0.65\columnwidth]{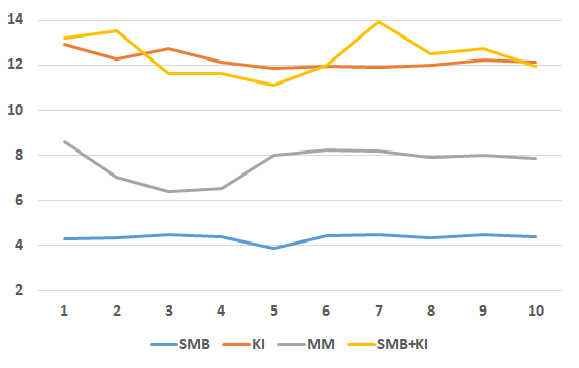} &
\includegraphics[width=0.65\columnwidth]{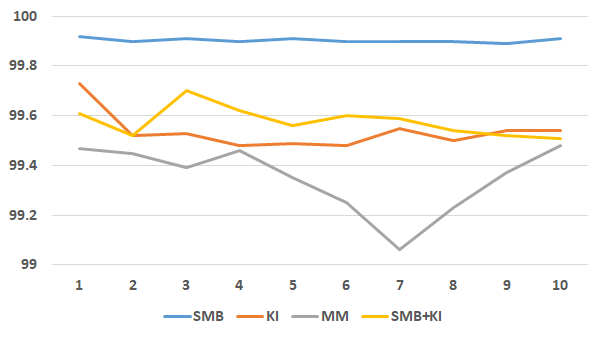}
\\
Density & Non-Linearity & Leniency \\
& & \\
\includegraphics[width=0.65\columnwidth]{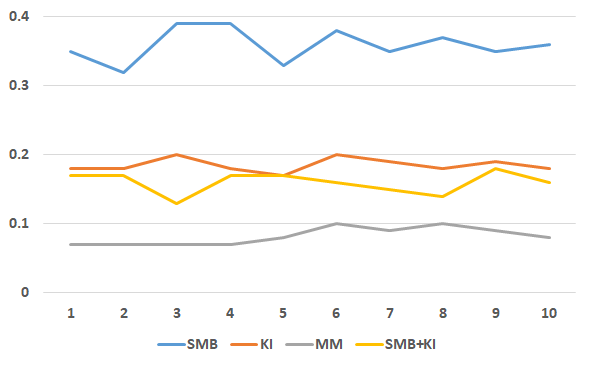} &
\includegraphics[width=0.65\columnwidth]{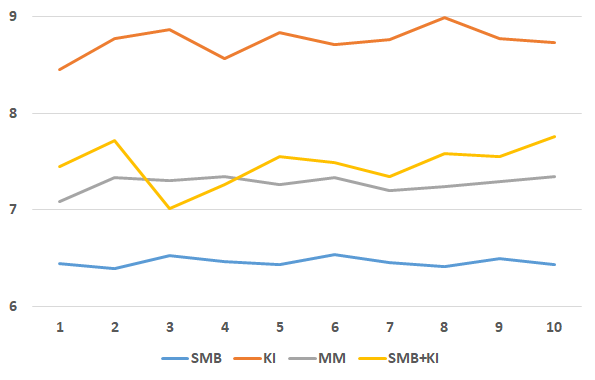} &
\includegraphics[width=0.65\columnwidth]{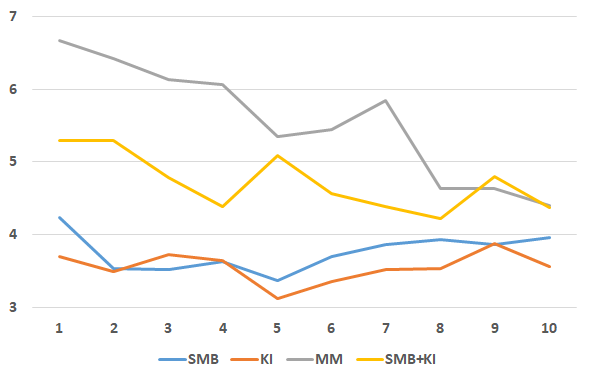}
\\
Interestingness & Path-Prop & Discontinuity \\
\end{tabular}
\caption{\label{XFIGUREprogression} \review{Per-segment metric values plotted for each grouping of 16 segments for MM and each grouping of 12 segments for the other games. x-axis values indicate 1st such grouping, 2nd such grouping etc. y-axis indicates average metric value for the corresponding group of segments.}}
\end{figure*}
}

\newcommand{\XFIGUREsmbcont}{
\begin{figure*}[h]
\centering
\begin{tabular}{c}
\raisebox{7pt}{\rotatebox{90}{\scriptsize{sequential}}}
\includegraphics[width=2\columnwidth]{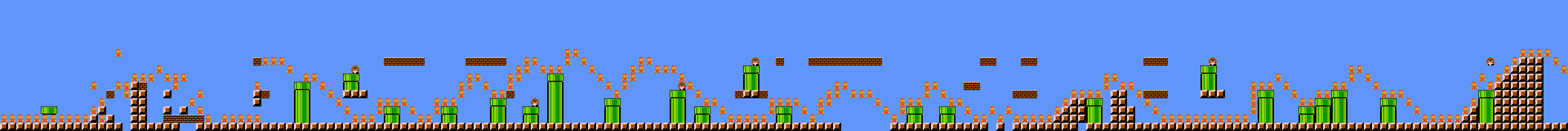}
\\
\raisebox{5pt}{\rotatebox{90}{\scriptsize{independent}}}
\includegraphics[width=2\columnwidth]{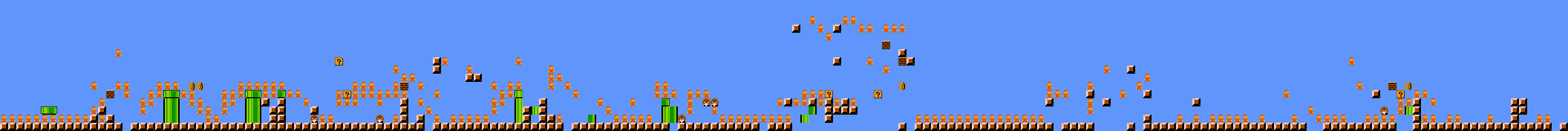}
\end{tabular}
\caption{\label{XFIGUREsmbcont} Example SMB levels generated using sequential (above) and independent (below) methods starting with the same randomly generated initial segment.}
\end{figure*}
}

\newcommand{\XFIGUREsmbfile}{
\begin{figure*}[h]
\centering
\begin{tabular}{c}
\raisebox{8pt}{\rotatebox{90}{\scriptsize{original}}}
\includegraphics[width=2\columnwidth]{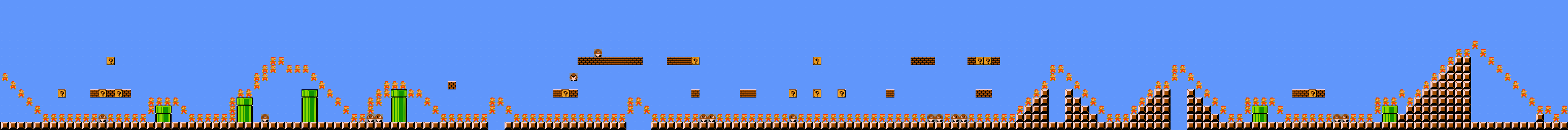}
\\
\raisebox{7pt}{\rotatebox{90}{\scriptsize{sequential}}}
\includegraphics[width=2\columnwidth]{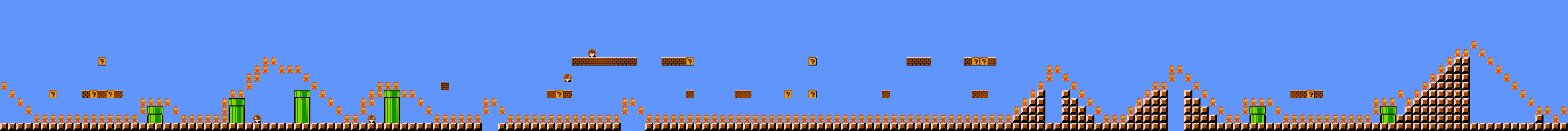}
\\
\raisebox{5pt}{\rotatebox{90}{\scriptsize{independent}}}
\includegraphics[width=2\columnwidth]{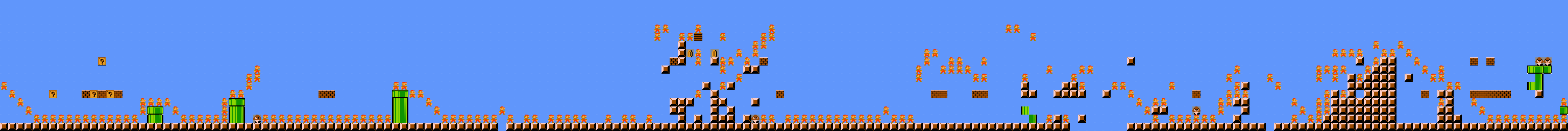}
\end{tabular}
\caption{\label{XFIGUREsmbfile} Original SMB level from the VGLC~\cite{summerville2016vglc} (top) and example levels generated with the initial segment of the original using the sequential (middle) and independent (below) methods.}
\end{figure*}
}

\newcommand{\XFIGUREki}{
\begin{figure}[h]
\setlength{\tabcolsep}{3pt}
\centering
\begin{tabular}{cccccc}
\includegraphics[width=0.155\columnwidth]{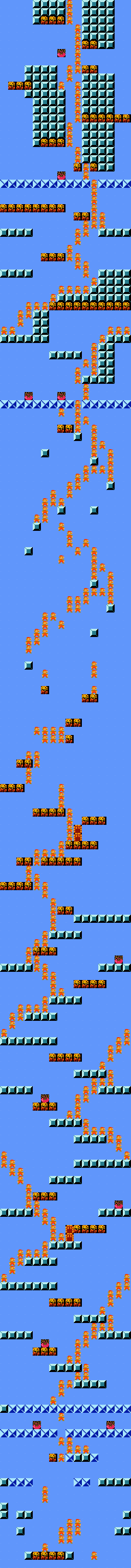} &
\includegraphics[width=0.155\columnwidth]{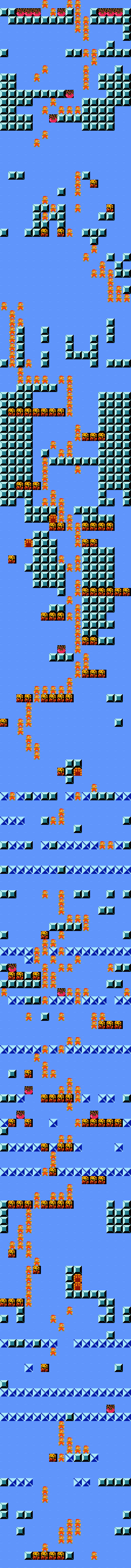} &
&
\includegraphics[width=0.155\columnwidth]{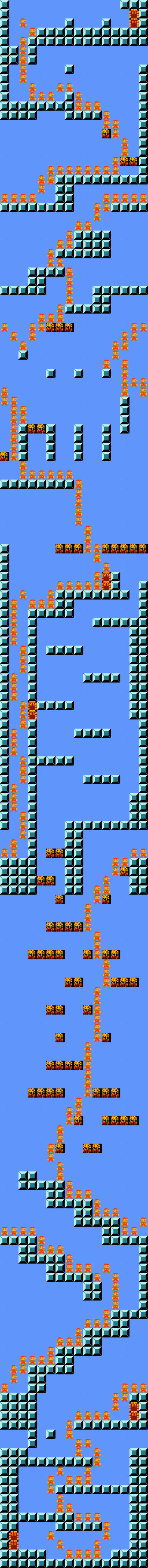} &
\includegraphics[width=0.155\columnwidth]{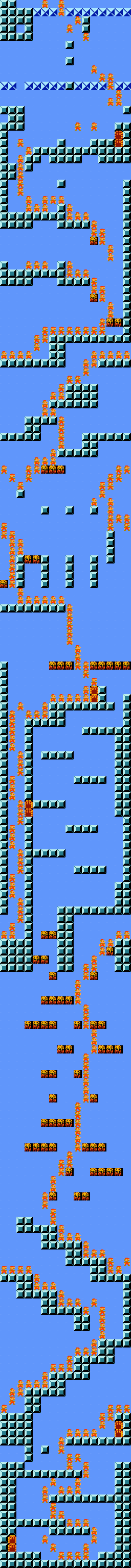} &
\includegraphics[width=0.155\columnwidth]{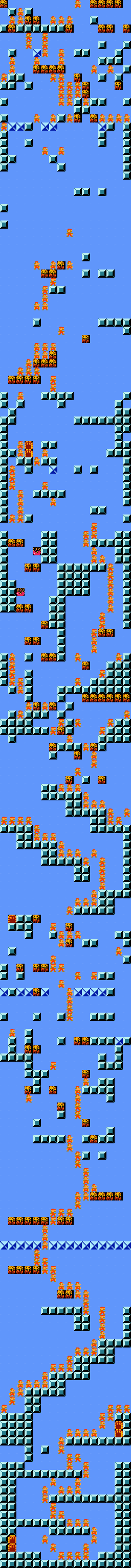} \\[-4pt]
\scriptsize{sequential} & \scriptsize{independent} & &
\scriptsize{original} & \scriptsize{sequential} & \scriptsize{independent} \\[2pt]
\multicolumn{2}{c}{(a)} &
&
\multicolumn{3}{c}{(b)} \\
\end{tabular}
\caption{\label{XFIGUREki} (a) Example KI levels generated using sequential (left) and independent (right) methods starting with same randomly generated initial segment. (b) Original KI level  from the VGLC~\cite{summerville2016vglc} (left) and example levels generated with initial segment of original using the sequential (middle) and independent (right) methods.}
\end{figure}
}

\newcommand{\XFIGUREmmcont}{
\begin{figure}[h]
\centering
\begin{tabular}{c}
\raisebox{45pt}{\rotatebox{90}{\scriptsize{sequential}}}
\includegraphics[width=0.95\columnwidth]{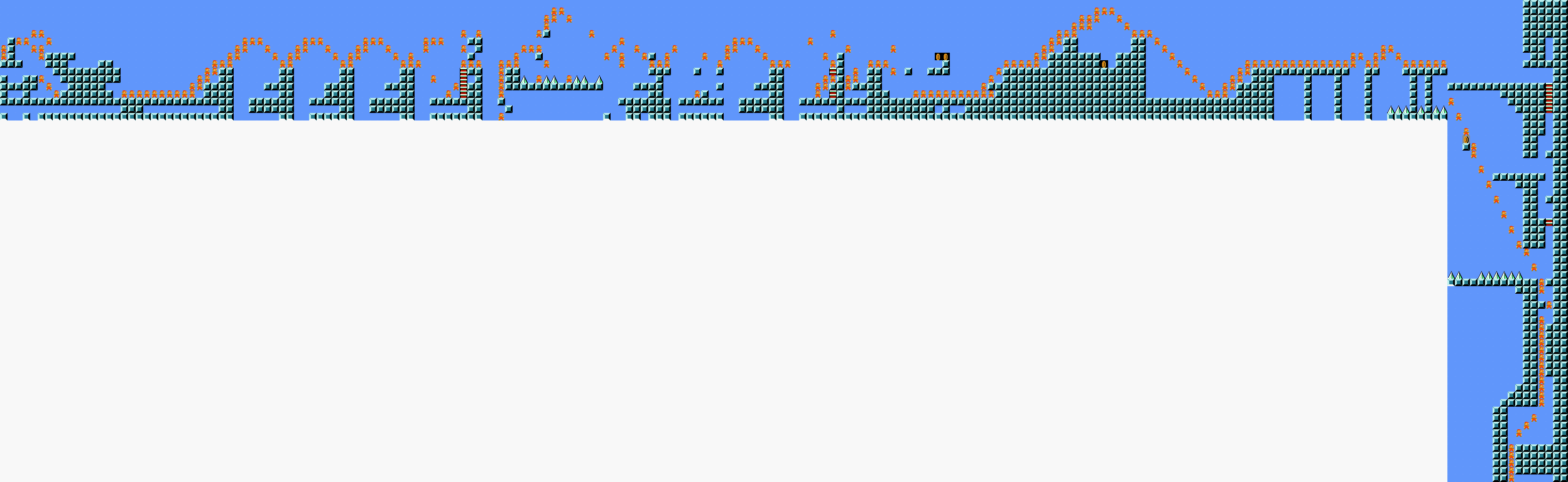}
\\
\raisebox{1pt}{\rotatebox{90}{\scriptsize{independent}}}
\includegraphics[width=0.95\columnwidth]{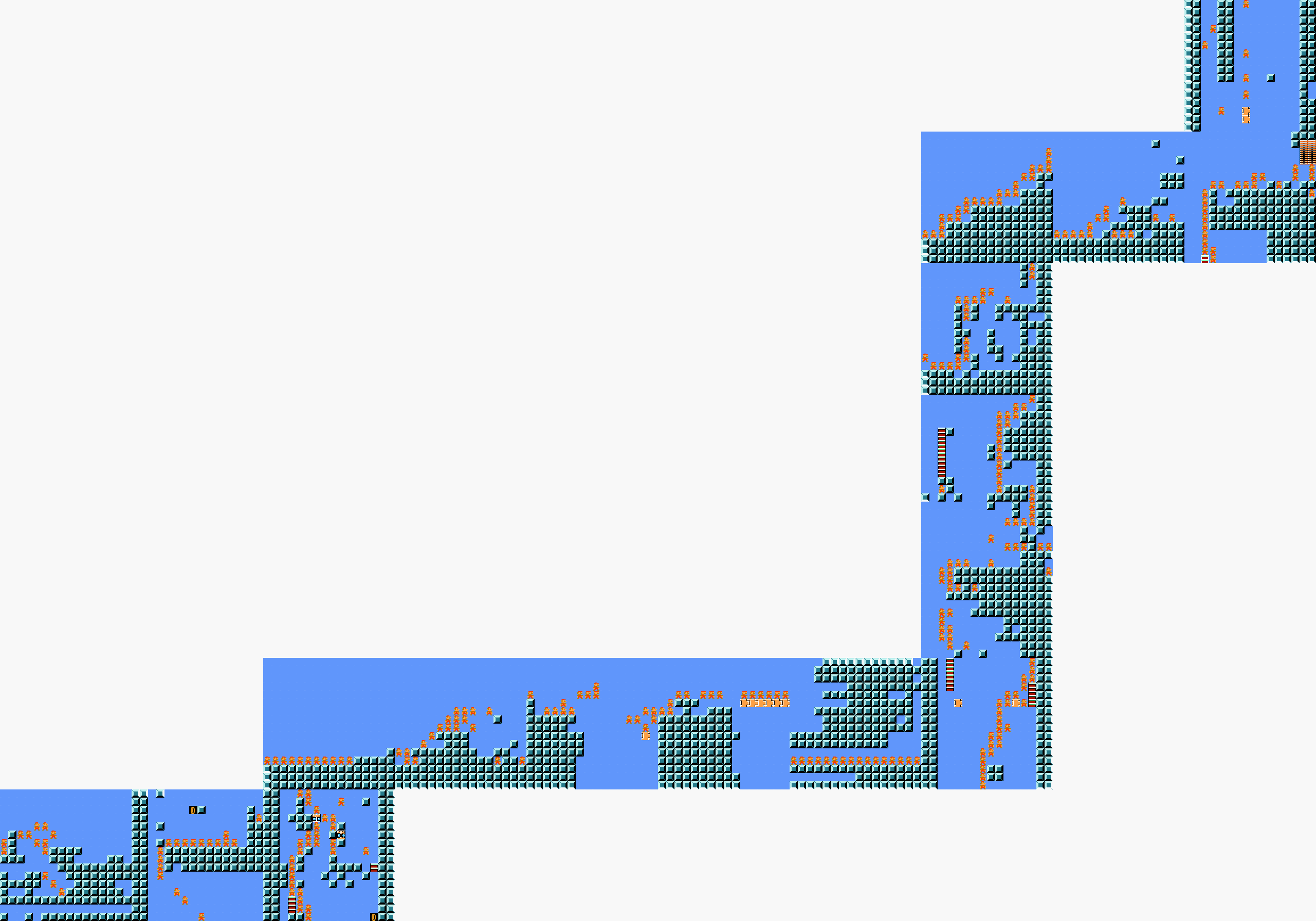}\\
\end{tabular}
\caption{\label{XFIGUREmmcont} Example MM levels generated using the sequential (top) and independent (bottom) methods with the same randomly generated initial segment.}
\end{figure}
}

\newcommand{\XFIGUREmmcontOLD}{
\begin{figure*}[h]
\centering
\begin{tabular}{cc}
\includegraphics[scale=0.125]{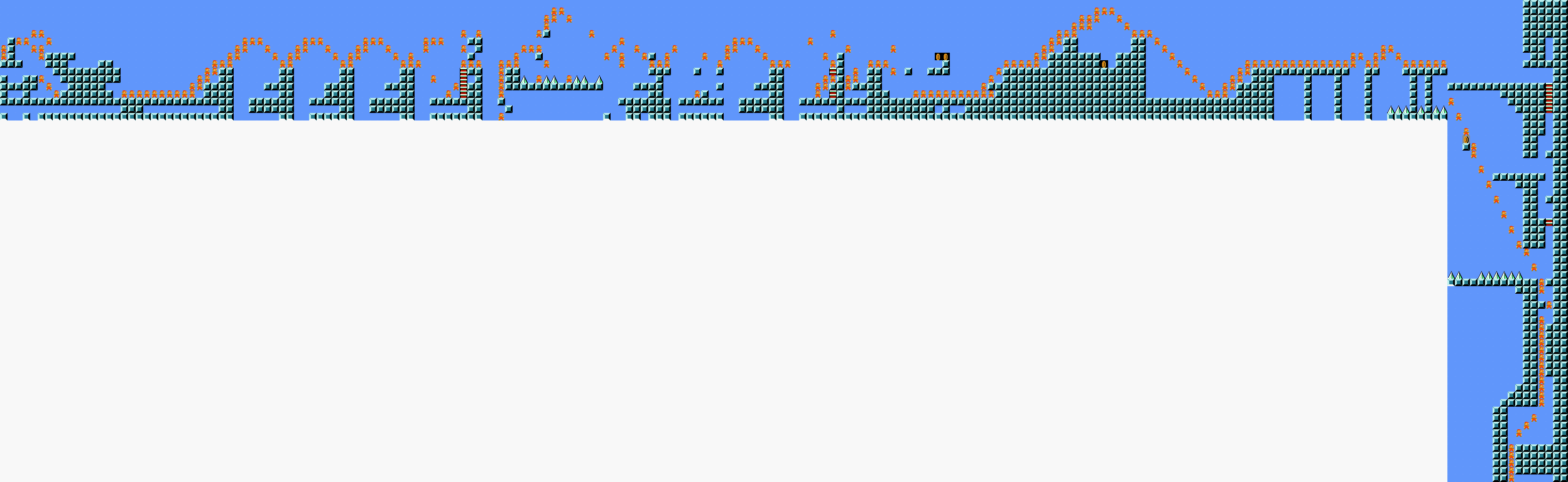}
&
\hspace{-0.2cm}
\includegraphics[scale=0.125]{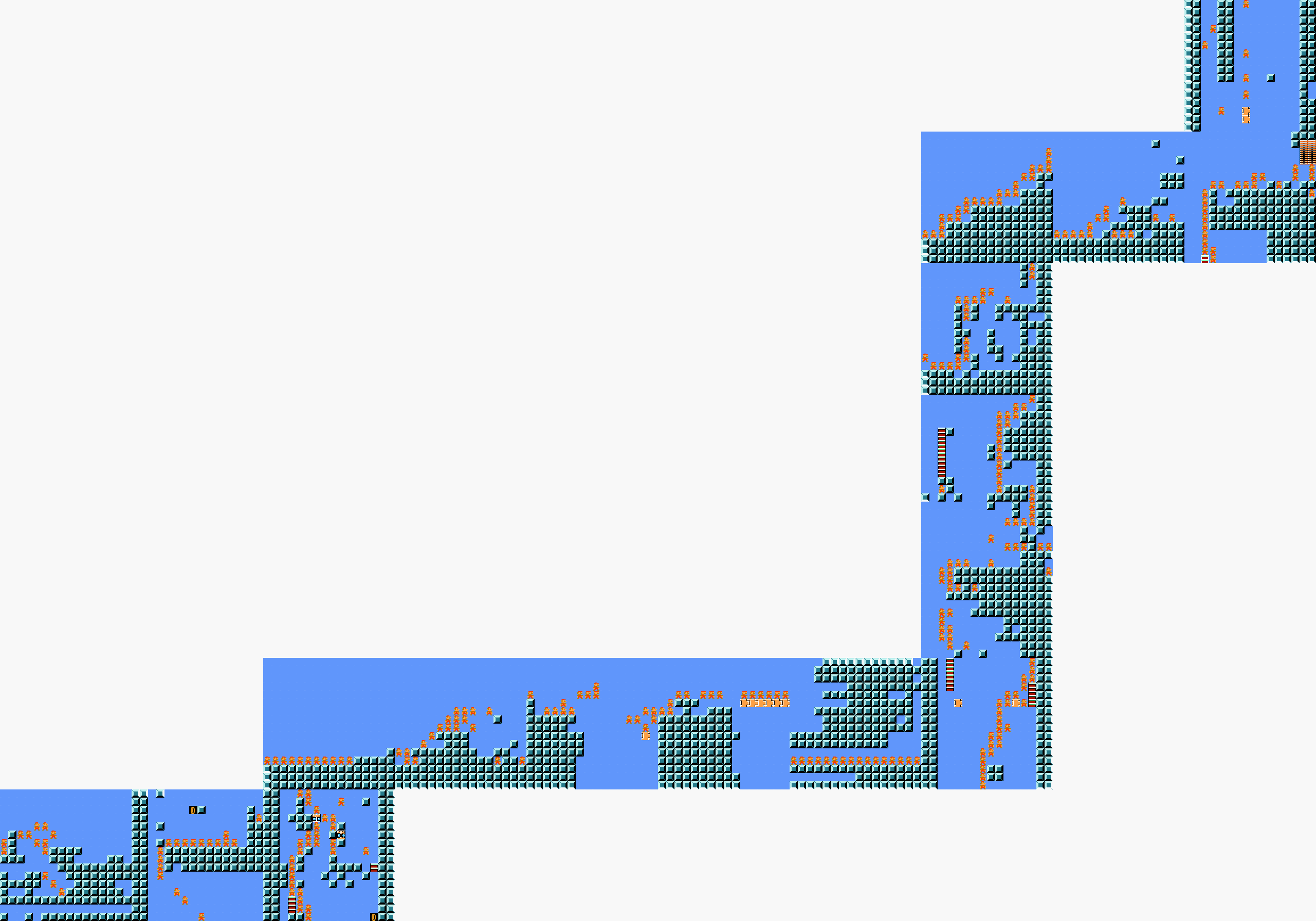}\\
\scriptsize{sequential} & \scriptsize{independent}
\end{tabular}
\caption{\label{XFIGUREmmcont} Example MM levels generated using the sequential (left) and independent (right) methods with the same randomly generated initial segment.}
\end{figure*}
}

\newcommand{\XFIGUREmmfile}{
\begin{figure*}[h]
\vspace*{0.2cm}
\centering
\begin{tabular}{c}
\raisebox{4pt}{\rotatebox{90}{\scriptsize{original}}}
\includegraphics[scale=0.11]{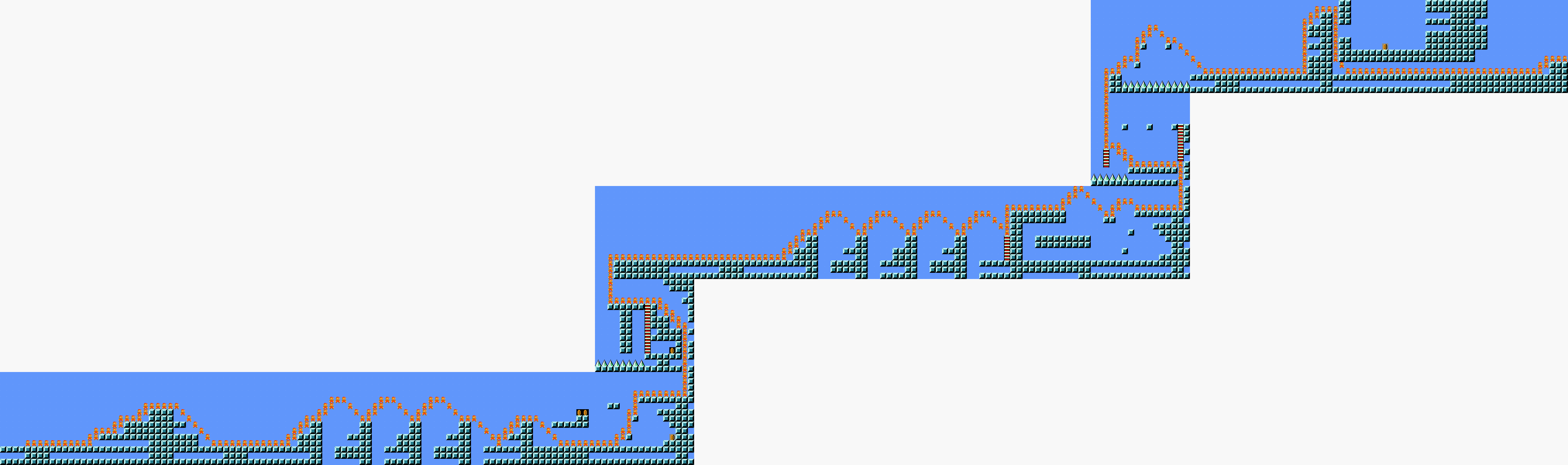}
\\
\raisebox{4pt}{\rotatebox{90}{\scriptsize{sequential}}}
\includegraphics[scale=0.22]{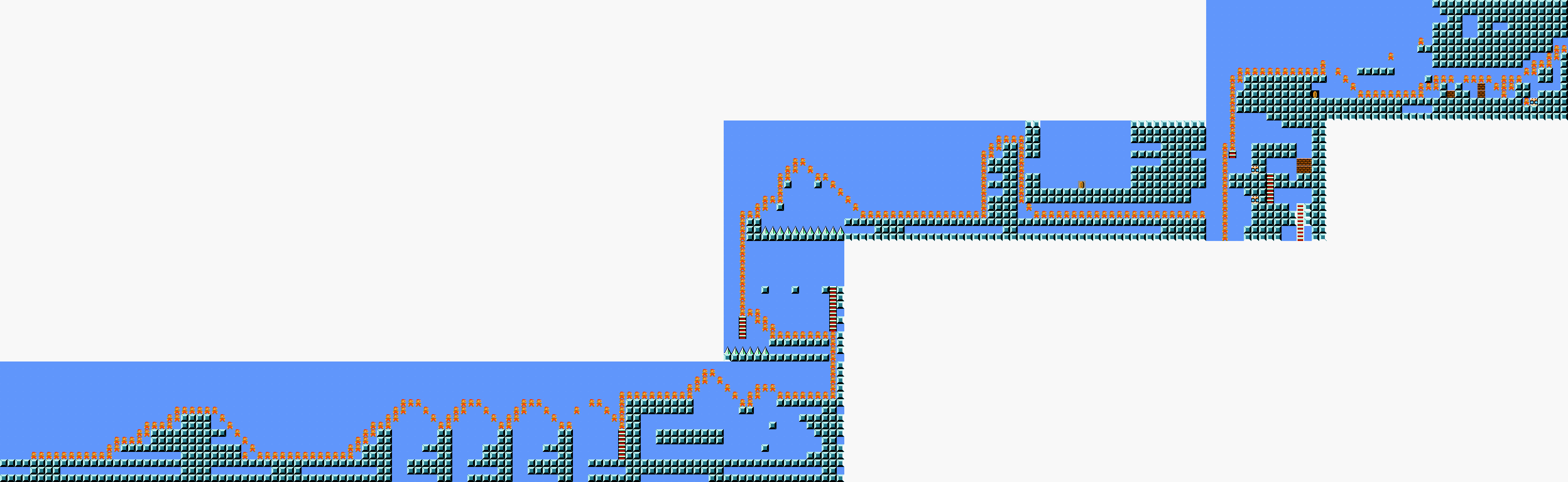}
\\
\raisebox{2pt}{\rotatebox{90}{\scriptsize{independent}}}
\includegraphics[scale=0.22]{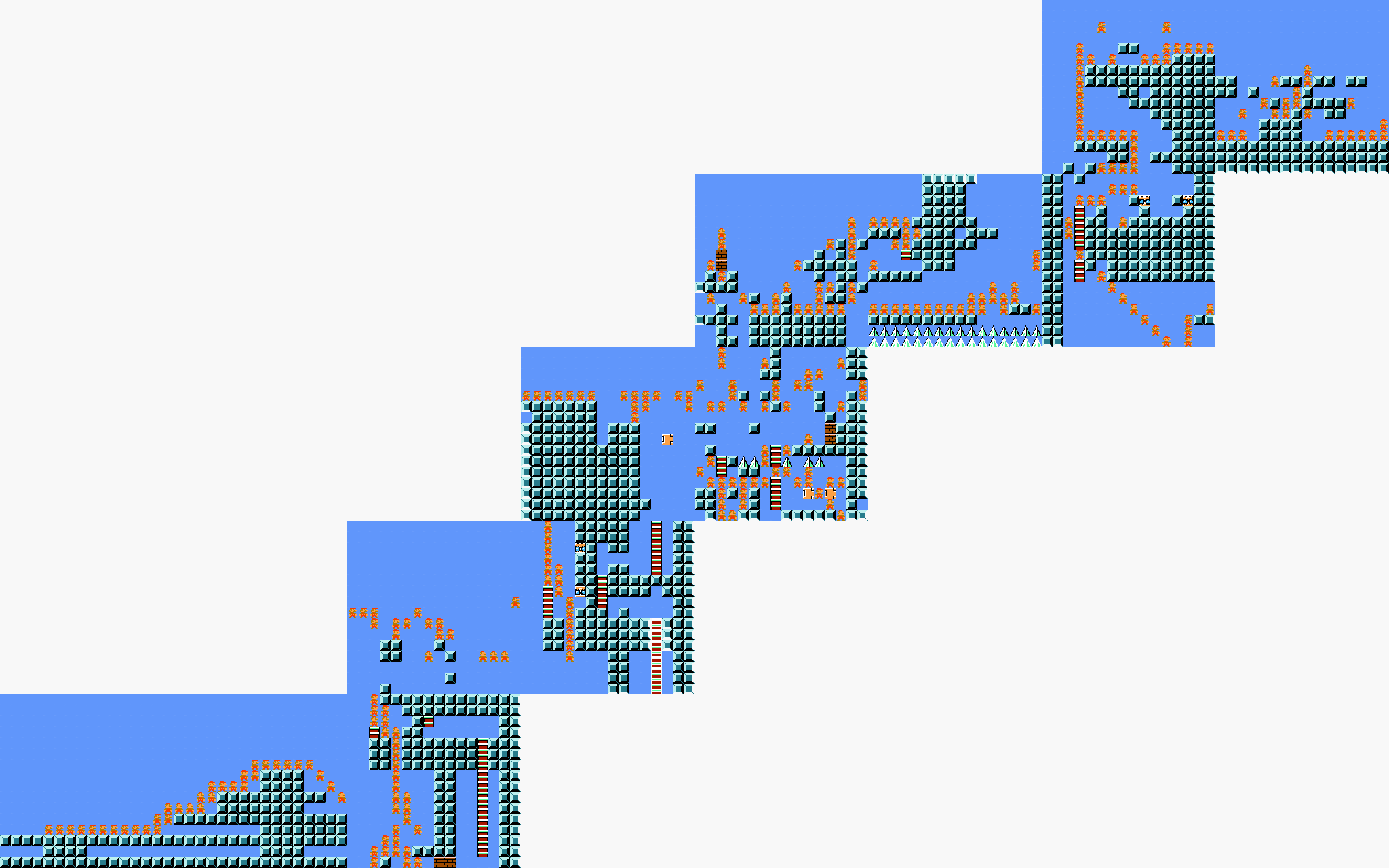}
\end{tabular}
\caption{\label{XFIGUREmmfile} Original MM level from the VGLC~\cite{summerville2016vglc} (top) and example levels generated with the initial segment of the original using the sequential (middle) and independent (bottom) methods.}
\end{figure*}
}

\newcommand{\XFIGUREbothcont}{
\begin{figure}[h]
\hspace*{-5pt}
\begin{tabular}{c}
\raisebox{1pt}{\rotatebox{90}{\scriptsize{sequential}}}
\includegraphics[width=0.96\columnwidth]{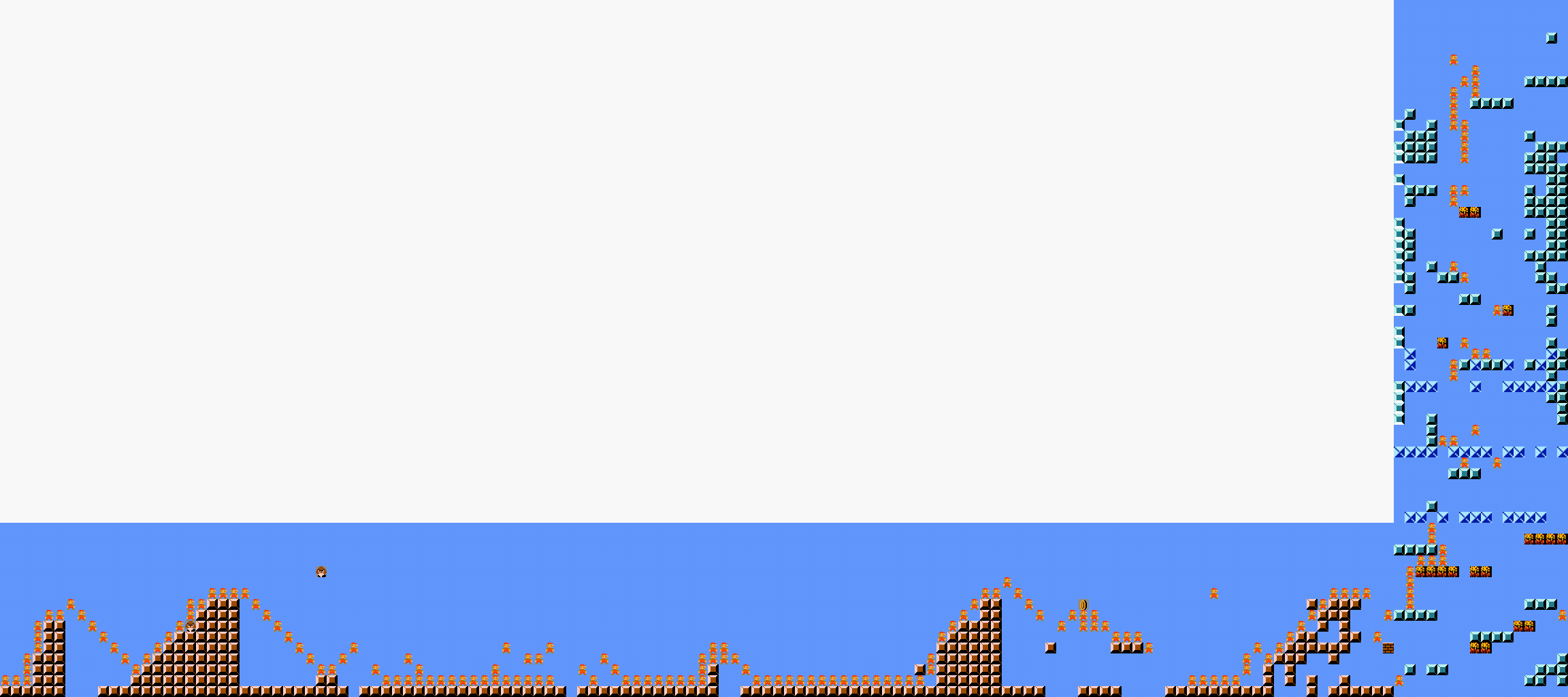}
\\
\raisebox{1pt}{\rotatebox{90}{\scriptsize{independent}}}
\includegraphics[width=0.96\columnwidth]{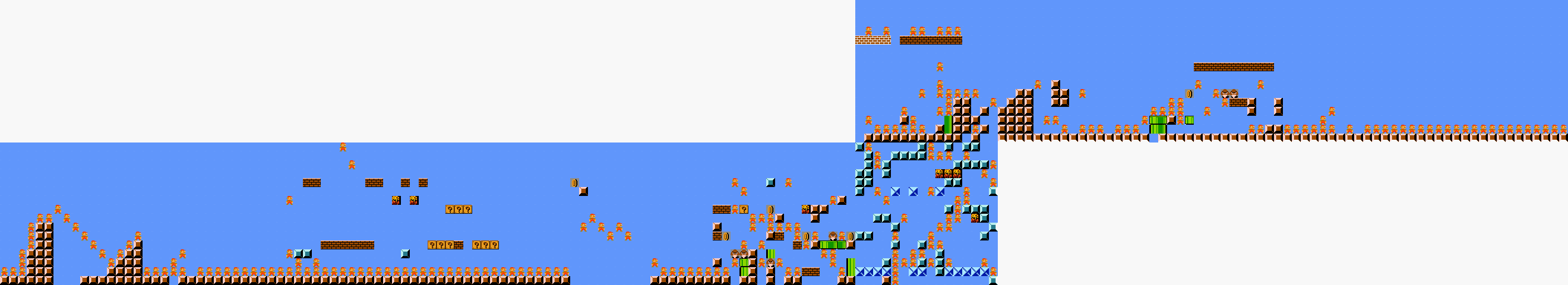}
\end{tabular}
\caption{\label{XFIGUREbothcont} Example blended SMB-KI levels generated using the sequential (above) and independent (below) methods starting with the same randomly generated initial segment.}
\end{figure}
}


\newcommand{\XTABLEcontinuity}{\begin{table}[t]
\begin{tabular}{|c|c|c|}
\hline
Game & Sequential & Independent \\
\hline
SMB & $3.86 \pm 2.28$ & $5.91 \pm 2.04$ \\
KI & $3.99 \pm 2.59$ & $7.37 \pm 1.99$ \\
MM & $6.54 \pm 2.63$ & $11.18 \pm 1.69$ \\
SMB-KI & $5.4 \pm 2.42$ & $9.84 \pm 1.76$ \\
\hline
\end{tabular}
\caption{\label{XTABLEcontinuity} Average per-segment \textit{Discontinuity} values along with standard deviation. A Wilcoxon Rank Sum Test showed differences to be significant with $p < .001$ in all cases.}
\end{table}
}

\newcommand{\XTABLEdirections}{\begin{table}[b]
\begin{tabular}{|c|c|c|}
\hline
Blend & SMB & KI \\
\hline
SMB-0 & 0.5 & 99.5 \\
SMB-25 & 4 & 96 \\
SMB-50 & 86.1 & 13.9 \\
SMB-75 & 85 & 15 \\
SMB-100 & 94.3 & 5.7 \\
Random Blend & 43.4 & 56.6 \\
\hline
\end{tabular}
\caption{\label{XTABLEdirections} Percentage of segments (out of $100 x 12 = 1200$) classified as SMB-like and KI-like using the directional classifier. }
\end{table}
}
\section{Introduction}
Procedural content generation via machine learning (PCGML) \cite{summerville2017procedural} refers to a subset of PCG techniques that use ML models for producing content. Several recent works \cite{volz2018evolving, sarkar2019blending, thakkar2019autoencoder, lucas2019tile} in this field have made use of latent variable models such as Generative Adversarial Networks (GANs) \cite{goodfellow2014generative} and Variational Autoencoders (VAEs) \cite{kingma2013autoencoding} for generating levels for platformers as well as dungeon crawlers. These models learn a continuous, latent representation of input game levels that can then be used to generate new levels via sampling, interpolating between points in this latent space, as well as by evolving latent vectors based on a given objective in order to generate levels with desired properties. While effective, these models work with fixed-size inputs and outputs, which limits the scope and coherence of levels that can be generated. Prior works produce levels by generating segments of levels independently and then stitching them together one after another. Though this can be satisfactory for platformers that proceed along a single direction or dungeon crawlers where gameplay can take place in discrete rooms, this approach would fail to generate playable, coherent levels for games where levels can proceed along multiple directions, both horizontally and vertically. Moreover, even for unidirectional platformers like \textit{Super Mario Bros.}, such methods are not ideal since randomly sampling successive level segments does not ensure that the current segment logically follows from previous ones.

In this work, we address these issues via a simple modification to the training procedure for such models, combined with the use of a classifier for segment placement. More specifically, we train VAEs on game level segments, but rather than training the model to reconstruct the current input segment, as is the norm, we train it to reconstruct the segment that follows it in the input game level. That is, decoding a segment's encoded latent representation yields not the segment itself but the segment immediately after it in the level's progression. Thus, starting from an initial segment, this approach enables the generation of arbitrarily long levels composed of segments that logically follow from one to the next, via an iterative loop of encoding and decoding. In addition to this modification, we also train a random forest classifier to determine whether the next segment should be placed above, below, to the left or to the right of the current segment. This VAE-classifier combination thus enables us to generate continuous, coherent platformer levels of desired length, progressing in multiple directions, while still working with fixed-sized segments. Moreover, such a model also enables the generation of not just segments that blend together multiple games as in prior work \cite{sarkar2019blending}, but entire blended levels without having to perform turn-based generation using multiple models as in \cite{sarkar2018blending}.

We demonstrate our approach for three different platformer games---\textit{Super Mario Bros.}, \textit{Kid Icarus}, and \textit{Mega Man}---as well as for the blended \textit{Super Mario Bros.--Kid Icarus} domain. The contribution of this work is a new generative approach that enables:
\vspace{-0.03cm}
\begin{enumerate}
    \item generation of more coherent platformer levels than possible using existing approaches;
    \item generation of levels for platformers like \textit{Mega Man} that progress in multiple directions; and
    \item generation of blended levels combining platformers progressing in different directions.
\end{enumerate}

\section{Related Work}
Methods for PCGML \cite{summerville2017procedural} attempt to generate new content by sampling models that have been trained on game data with the hope of producing content that is novel but also captures the patterns and properties of the games used for training. 
While numerous ML techniques (including autoencoders \cite{jain_autoencoders_2016}, LSTMs \cite{summerville2016mariostring}, Markov models \cite{dahlskog2014linear, snodgrass2017learning} and Bayes Nets \cite{guzdial2016learning, summerville2016learning}) have been used for generating levels, more recent ML advances such as latent variable models like Generative Adversarial Networks (GANs) and Variational Autoencoders (VAEs) are also being increasingly used for generating game content. Volz et al. \cite{volz2018evolving} used a GAN to generate Mario levels and demonstrated the feasibility of using the learned latent space for evolving levels with desired properties. In a similar vein, Sarkar et al. \cite{sarkar2019blending} used VAEs to generate level segments that blended the properties of \textit{Super Mario Bros.} and \textit{Kid Icarus} and exhibited the utility of the VAE's latent space for evolving blended content with desired characteristics. In addition to Mario, GANs have also been used to generate \textit{Doom} levels \cite{giacomello2018doom} and \textit{Legend of Zelda} dungeons \cite{gutierrez2020generative} while VAEs have been used to generate \textit{Lode Runner} levels \cite{thakkar2019autoencoder}. However, due to the nature of GANs and VAEs, all such approaches are required to work with fixed-size inputs and outputs and are thus forced in many cases to train on and generate segments of levels rather than levels themselves, since using entire levels as the fixed-size inputs would normally lead to an insufficient amount of training data. In GAN-based Mario generation \cite{volz2018evolving}, authors prescribe generating whole levels by simply stitching together generated segments. While this works due to the simple nature of Mario only progressing along one direction, it is not ideal since segments are generated from latent space vectors, each of which independently encodes a single segment, and thus there is no guarantee that successively generated segments follow each other optimally. Additionally, such an approach would not work for more complex platformers which can progress in multiple directions or to blend together segments from different games as in prior work \cite{sarkar2019blending} which ignores generating whole levels and restricts generation to segments alone. Thus, in this paper, we attempt to address the problem of generating continuous, whole levels for a variety of platformers while contending with the fixed-sized limitations of latent variable models. Recently, Gutierrez et al. \cite{gutierrez2020generative} also addressed this issue of generating entire levels using fixed-sized inputs and outputs. In their work, they use GANs to generate fixed-size rooms and connect them using a graph grammar to form a dungeon. Our approach differs in that it uses VAEs, works with platformers and conditions the generation of a segment on the previous segment. 

Another recent trend in PCGML research has been to focus on more creative uses of ML \cite{guzdial2018combinatorial} for generating game content. Such works try to move beyond generating levels for an existing game or domain in order to enable PCG techniques such as game blending, domain transfer and automated game generation. These methods fall under combinational creativity \cite{boden2004creative}, the branch of creativity in which existing concepts and domains are combined to generate novel ones. Previous such methods include Snodgrass and Onta{\~n}{\'o}n's domain transfer work \cite{snodgrass2016approach}, Guzdial and Riedl's conceptual expansion technique \cite{guzdial2018automated} that generates new games by combining levels and rules of existing games using Bayes nets and game graphs, Snodgrass and Sarkar's \cite{snodgrass2020multi} hybrid model combining binary space partitioning and VAEs to generate blended levels across multiple platformer games using a sketch representation and Sarkar et al.'s \cite{sarkar2019blending} use of VAEs for blending level segments of \textit{Super Mario Bros.} and \textit{Kid Icarus}. Our work builds directly on the latter by enabling generation of entire blended levels rather than just segments and additionally generating blended levels that are mostly traversable unlike the blended segments in the previous work.

\section{Method}

We describe the game level data used in this work as well as the two main parts to the generative pipeline---the VAE for segment generation and the random forest classifier for segment placement classification.

\subsection{Level Data}
We demonstrate our approach using levels from 3 classic NES platformers---\textit{Super Mario Bros.} \cite{supermario:nes}, \textit{Kid Icarus} \cite{kidicarus} and \textit{Mega Man} \cite{megaman}, henceforth referred to as SMB, KI and MM respectively. Additionally, to test blending, we also used a combined SMB-KI domain. All levels were taken from the Video Game Level Corpus (VGLC) \cite{summerville2016vglc}. These levels use a tile-based text representation and are annotated with the path of an $A^*$ agent tuned using the jump arcs of the game \cite{summerville2017mechanics}. This allows a trained model to produce such paths in the generated levels and thus help make them playable. While SMB and KI levels progress exclusively left-to-right and bottom-to-top respectively, MM levels progress left-to-right as well as both bottom-to-top and top-to-bottom. Thus, past approaches for VAE and GAN-based level generation would not be able to reliably generate MM levels since consecutive randomly generated segments could be oriented incompatibly, which we address by conditioning on the previous segment in our approach. Similarly, levels in a blended SMB-KI domain would be expected to progress both to the right and to the top and would necessitate a similar modification in order to be amenable for generation. For all games, we train our models on 16x16 segments produced by sliding a window of that size horizontally and vertically across levels as appropriate given the orientation of the game. Horizontal segments of SMB and MM are originally 14 and 15 rows high respectively so we pad them with additional row(s) of all background tiles to have a uniform height of 16. This gave us 2458 segments for SMB, 1046 segments for KI and 1572 segments for MM. For the blended SMB-KI domain, to better balance the number of segments, we doubled the KI segments to end up with 2092. Additionally, for levels in the blended domain, we used the original tiles from each game's VGLC representation except for using a common tile for background and path. For all domains, paths are represented using a Mario character sprite.

\XFIGURErecon

\subsection{Sequential Segment Generation using VAEs}
To build our models, we trained a VAE on each of SMB, KI and MM as well as on the blended SMB-KI domain. VAEs \cite{kingma2013autoencoding} learn continuous, latent representations of data, and consist of an encoder network which learns to map data to a lower-dimensional vector in the latent space and a decoder network which learns to reconstruct the original data from this latent vector. This is achieved by training via minimizing a loss function that consists of two terms: 1) the reconstruction error and 2) the Kullback-Leibler (KL) divergence---a statistical measure of the similarity between two probability distributions. Minimizing the former reduces the error between inputs and reconstructed outputs produced by the decoder while minimizing the KL divergence between the latent distribution and a known prior (typically a Gaussian) enforces the latent space to model a continuous, informative distribution. We modify the computing of the reconstruction error to enable our approach.

Our input consists of 16x16 segments. Typically, the reconstruction error would be computed between the segment output by the decoder and the corresponding segment that was passed through the encoder. Instead, in our approach, we compute this error between the decoder output and the segment that \textit{follows} the corresponding segment that was encoded. This simple modification enables the VAE to learn a sequential model of segment generation where the encoder maps a given segment into a latent vector but the decoder maps that latent vector into the segment that would sequentially follow the original segment in a level. This is depicted in Figure \ref{XFIGURErecon}. The algorithm shown below thus enables generation of a sequence of segments that follow a logical gameplay progression and hence can be combined into a coherent level.

\begin{algorithm}
\caption{GenerateLevel(init\_segment, n)}
\label{alg1}
\begin{algorithmic}
\STATE Initialize $level$ to $init\_segment$
\STATE $num\_segments = 1$
\STATE $segment = init\_segment$
\WHILE{$num\_segments \leq n$}
\STATE $z \leftarrow Encoder(segment)$
\STATE $segment \leftarrow Decoder(z)$
\STATE Add $segment$ to $level$
\STATE $num\_segments \mathrel{+}= 1$
\ENDWHILE
\RETURN $level$
\end{algorithmic}
\end{algorithm}

All models were trained using PyTorch \cite{paszke2017automatic} and used the same architecture. Encoders and decoders each consisted of 4 linear layers with ReLU activation. The decoder output was further passed through a sigmoid layer. All models used a 128-dimensional latent space and were trained for 10000 epochs with the Adam optimizer and a learning rate of 0.001 decayed by 0.1 every 2500 epochs. Additionally, to aid in training, the weight of the KL-divergence term in the variational loss was annealed linearly from 0.0 to 1.0 over the first 2500 epochs.

\subsection{Placement Classification}
While the above modification gives us a sequential segment-based level generation model, it still only improves upon existing generation approaches along one direction. To generate levels that can dynamically progress along any direction, we need to determine where to place a generated segment in relation to the previous segment. For this purpose, for each domain, we train a random forest classifier on its segments. Here the inputs are the 16x16 segments and each label is the direction where the next segment appears in the original level. Thus, given a generated segment, the classifier determines in which direction (up, down, left or right) it should be placed with respect to the previous segment. We trained the classifiers using a 70\%-30\% train-test split for each domain, obtaining accuracies of 100\% on each of SMB, KI and SMB-KI and 98.73\% for MM. For the MM training set, we oversampled the vertically progressing segments till we had an equal number of segments in all directions to account for class imbalance. Combining the classifier with the VAE, gives us the following generative pipeline.

\begin{algorithm}
\caption{GenerateLevelWithDirs(init\_segment, n)}
\label{alg2}
\begin{algorithmic}
\STATE $level \leftarrow $ GenerateLevel(init\_segment, n)
\STATE $level\_with\_dirs \leftarrow \emptyset$
\FOR{$segment$ in $level$}
\STATE $dir \leftarrow Classifier(segment)$
\STATE Add $(segment, dir)$ to $level\_with\_dirs$
\ENDFOR
\RETURN $level\_with\_dirs$
\end{algorithmic}
\end{algorithm}

Note that for the domains we used, only MM and the SMB-KI blend require a classifier. All levels in SMB and KI progress in one direction (right and up respectively) so in these cases, it is sufficient to simply place the generated segments one after another to get a coherent level. However for generality, we still used the classifier for every domain in all of our evaluations. A potential pitfall of using the classifier is dealing with segments that are incorrectly classified. The most common form of mis-classification that could disrupt the progression of a level is along an individual axis (i.e. segments that should be followed in the downward direction misclassified as upward and vice-versa; segments that should be followed to the right being misclassified as left and vice-versa). These are common because often times such segments taken individually could make sense in either direction. Segments that proceed upward and downward in MM for example, often share similar structures. Similarly, an individual segment in SMB taken in a vacuum could equally progress to the left or to the right. In such cases, it is the progression of the level generated up to that segment that determines the correct direction for the next segment rather than the segment itself. To account for this potential issue, we prevent the classifier from predicting the next direction to be the direction that connects the newly generated segment to the previous one i.e. the direction that if the newly generated segment were to be placed using it, would overwrite the previous segment. This is done by using the classifier prediction with the second highest likelihood if the prediction with the highest is the direction to avoid.

\section{Results}
We tested our approach using a three-part evaluation, focusing on 1) the continuous nature of generated levels compared to past methods, 2) properties of generated blended levels and 3) the quality of generated levels that are arbitrarily long. We describe each part in the following sections.

\subsection{Discontinuity}
To test if levels generated using our methods have a better sense of progression than past methods, we introduced a \textit{Discontinuity} metric. We define this as the absolute distance between path tiles along the adjoining edge of two successive segments. That is, if two segments are connected horizontally, we compute the displacement between the path tiles on the columns at the edge connecting the two segments. If either column does not have a path tile, the metric returns 16 by default since the maximum height of a column is 16 and thus the path tiles can be at most 15 tiles from each other. For segments connected vertically, this is similarly computed using the rows at the edge of the two segments. Thus, lower the value, the more continuous the path is from one segment to the next. The reasoning for this metric is that levels with a better sense of progression would have a more continuous path through its segments rather than a path with a lot of displacement between where it ends for one segment and where it begins for the next. While by no means a perfect measure of progression through a level, it nevertheless gives a sense of the nature of the path through a level and is thus suitable for comparison with past generative methods. For our computations, we ignored the fact that KI levels wrap around horizontally.

For evaluation, we generated 100 levels each for SMB, KI, MM and SMB-KI using two methods: 1) using the generative loop described in Algorithm \ref{alg2}, where segments are generated conditioned on the previous, and 2) stitching together segments generated independently of each other from randomly sampled latent vectors, which is how whole levels are generated using past methods. For the rest of the paper, we refer to these methods as \textit{sequential} and \textit{independent} respectively. Each generated level consisted of 12 segments for SMB and KI and 16 for MM since those were respectively the average number of 16x16 segments in levels from the original games. For combined SMB-KI, generated levels consisted of 12 segments. Results are shown in Table \ref{XTABLEcontinuity}.

For all games, the sequential method led to significantly lower \textit{Discontinuity} values than the independent method, thus suggesting that levels generated using the former have more continuous paths through their segments. Note that the differences are greater for MM and SMB-KI since as we have discussed, segments generated using the independent method for these two often leads to levels that are not traversable where as for SMB and KI, the independently generated levels are often playable even if not as continuous as sequentially generated ones.

Examples of sequential and independently generated levels for SMB are in Figures \ref{XFIGUREsmbcont} and
\ref{XFIGUREsmbfile}, for KI in Figure \ref{XFIGUREki}, for MM in Figures \ref{XFIGUREmmcont} and \ref{XFIGUREmmfile}, and for SMB-KI in Figure \ref{XFIGUREbothcont}. Based on visual inspection, the sequential method generates levels with a more continuous flow from segment to segment than the independent method. While expected, it is worth noting that using the sequential method, starting generation with the initial segment of an original level produces a level very similar to the original, as seen in Figures \ref{XFIGUREsmbfile}, \ref{XFIGUREki}(b) and \ref{XFIGUREmmfile}, while the independent method of course does not do so since segments are generated independently. A byproduct of independent segment generation is that independently generated levels seem to be more diverse, less predictable and result in more directional changes for games that progress in multiple directions. In the right context, these are desirable features and thus it would be interesting in the future to look at methods capable of trading off between the better playability and continuity of the sequential method with the increased variety of the independent approach.

\XTABLEcontinuity

\XFIGUREblendingsepaltsub

\XTABLEdirections

\XFIGUREprogressionsub

\subsection{Blending}
The ability to generate levels progressing in multiple directions also enables us to move beyond generating segments that blend games progressing in different directions to entire levels that do so. To test blending, we generated 100 12-segment blended SMB-KI levels using a VAE trained on segments from both games as described previously. We generated 6 sets of 100 such levels with each set differing in the choice of initial segment. The first set used initial segments sampled randomly from the SMB-KI latent space. The remaining five used initial segments obtained by interpolating between latent vectors corresponding to actual SMB and KI segments at intervals of 25\%. Thus we obtained 5 such sets labeled as SMB-0, SMB-25, SMB-50, SMB-75 and SMB-100, indicating the distance interpolated from the KI segment to the SMB segment. We compared the blended levels with the original SMB and KI levels using the following tile-based segment-level metrics:

\begin{itemize}
    \item \textit{Density}: the proportion of a segment occupied by tiles that the player can stand on such as blocks, ground and platforms
    \item \textit{Non-Linearity}: a measure of how a segment's topology fits to a line, calculated by computing the mean square error of running linear regression on the topmost point of columns in a segment. Zero value indicates perfect linearity
    \item \textit{Leniency}: the proportion of a segment that is not occupied by any enemy or hazard tiles
    \item \textit{Interestingness}: the proportion of a segment occupied by interactable items such as collectables and powerups
    \item \textit{Path-Prop}: the proportion of a segment occupied by path tiles
\end{itemize}

\XFIGUREsmbcont

\XFIGUREsmbfile

\noindent Results are given in Figure \ref{XFIGUREblendingsepaltsub} and show that the values for blended levels fall mostly between those for SMB and KI, suggesting that the properties of the generated levels do blend those of the two original games, especially in terms of \textit{Density}, \textit{Path-Prop} and \textit{Nonlinearity}. Values for \textit{Leniency} and \textit{Interestingness} for blended levels are between those for SMB and KI as well but not in the expected pattern and speaks to generated levels having fewer enemies, hazards and collectible items than the originals.

Additionally, we also looked at the proportion of SMB-like and KI-like segments that were generated for the different sets of blends. Since we know that SMB and KI levels progress exclusively to the right and upward respectively, we can use our directional classifier as a proxy for this purpose. That is, blended segments classified to have the next segment to the right can be deemed to be more SMB-like while those classified to have the next segment be to the top can be deemed KI-like. Results for this are shown in Table \ref{XTABLEdirections}. We see that levels generated from random latent vectors perform the most amount of blending while those generated from interpolated vectors heavily favor the game with the higher proportion in the interpolation, mostly ignoring the actual proportion itself. Thus, while this approach is successful in generating blended levels, in the future, we would like to augment it such that the nature and amount of blending is more controllable.

\subsection{Progression}
For our final evaluation, we wanted to demonstrate the ability of our approach to generate arbitrarily long levels without the quality or characteristics of later segments deteriorating. We generated 100 levels of 120 segments for each of SMB, KI and SMB-KI and 100 levels of 160 segments for MM (i.e. levels approximately 10 times the size of an average level) and computed the average \textit{Discontinuity} and tile-based metrics defined above per segment for each of the 10 sub-levels (i.e. each set of 16 segments in MM and each set of 12 segments in the others). Results for this are shown in Figure \ref{XFIGUREprogressionsub}. Ideally, we would like to see little variation in terms of metric values for all 10 sets of segments and this indeed bears out in the results for all metrics and games except for \textit{Discontinuity} for MM which actually seems to get lower as more segments are generated. A possibility for this is that the MM model might be falling into a pattern of generating similar (or the same) segments over and over again causing little variation in path and hence low \textit{Discontinuity}.  This ties into the broader problem of VAEs suffering from \textit{posterior collapse} \cite{lucas2019dont,lucas2019understanding,razavi2019preventing} where the decoder learns to reconstruct data while ignoring a subset (or in the worst case, all) of the latent dimensions, resulting in an uninformative latent space. Though we trained our models using KL-annealing \cite{bowman2016generating, fu2019cyclical} to specifically account for this, and we do not encounter it for regular-sized levels, it is possible that the problem manifests itself when trying to generate larger levels and needs to be studied more thoroughly in the future.

\XFIGUREki

\XFIGUREbothcont

\XFIGUREmmcont

\XFIGUREmmfile

\section{Conclusion and Future Work}
In this work, we presented a novel PCGML approach that combines the use of a variational autoencoder and a random forest classifier to produce a model for sequential platformer level generation. Our results demonstrate that this enables generation of more coherent platformer levels than past approaches, generation of platformer levels that progress in multiple directions, blending of levels from games that progress differently and generation of levels that are arbitrarily long without suffering from a loss of quality. There are several considerations for future work.

While we evaluated our model in terms of continuity of generation and blending, we did not specifically evaluate the classifier on generated segments, mainly in part due to generated segments not having a ground truth to test against. In the future, we could compare classifier-based segment placement with other placement strategies, optimizing for metrics such as continuity and playability. 

Our approach demonstrated the feasibility of generating levels for multi-directional platformers such as \textit{Mega Man} but there is massive room for improvement in terms of quality of models and generated levels, reliability of generation and controllability, all of which should be investigated in future work. Moreover, none of the games we used had levels progressing from right-to-left. While our classifier was trained to work with all four directions in mind and should work as is for right-to-left progression, this needs to be empirically validated in the future.

Additionally, while our method demonstrably allows for generation of traversable, blended levels, the current approach does not have any direct means of controlling the blend proportions. It is possible to use some of the related latent variable evolution strategies from \cite{sarkar2019blending} but this needs to be empirically tested in future work. To this end, we could use conditional variants of the VAE \cite{mirza2014conditional}. Such models can explicitly condition the generation of segments on properties such as direction and game type and thereby allow designers to generate segments with greater control.

Finally, we intend to test this approach with more games and more diverse blends composed of more than two games. The ability to generate traversable blended levels opens up the possibility of generating new mechanics which combined with these newly blended levels could form the foundation for generating entire blended games in the future.


\bibliographystyle{ACM-Reference-Format}
\bibliography{refs-custom}

\end{document}